\documentclass{article}
\PassOptionsToPackage{numbers}{natbib}

\usepackage[main,final]{neurips_2026}

\usepackage[utf8]{inputenc} % allow utf-8 input
\usepackage[T1]{fontenc}    % use 8-bit T1 fonts
\usepackage{hyperref}       % hyperlinks
\usepackage{url}            % simple URL typesetting
\usepackage{booktabs}       % professional-quality tables
\usepackage{amsfonts}       % blackboard math symbols
\usepackage{nicefrac}       % compact symbols for 1/2, etc.
\usepackage{microtype}      % microtypography
\usepackage{xcolor}         % colors

\usepackage{enumitem}
\usepackage{graphicx}
\usepackage{booktabs}
\usepackage{multirow}
\usepackage{wrapfig}
\usepackage{graphicx}
\usepackage{pifont}
\usepackage{makecell}
\usepackage[utf8]{inputenc}
\usepackage{booktabs} % For professional-looking table lines
\usepackage{fancyvrb} % The key package for verbatim text in a cell
\usepackage{booktabs}
\usepackage{tabularx}
\usepackage{verbatimbox}
\usepackage{xltabular}
\usepackage{listings}
\usepackage[most]{tcolorbox} % 强烈推荐加 [most]，支持更多样式
\usepackage{xcolor}

\usepackage{enumitem}
\usepackage{graphicx}
\usepackage{booktabs}
\usepackage{multirow}
\usepackage{wrapfig}
\usepackage{graphicx}
\usepackage{pifont}
\usepackage{makecell}
\usepackage[utf8]{inputenc}
\usepackage{booktabs} % For professional-looking table lines
\usepackage{fancyvrb} % The key package for verbatim text in a cell
\usepackage{booktabs}
\usepackage{tabularx}
\usepackage{verbatimbox}
\usepackage{xltabular}
\usepackage{listings}
\usepackage{hyperref}

\usepackage{amsmath}
\usepackage{amssymb}
\usepackage{mathtools}
\usepackage{amsthm}
\usepackage{caption}
\usepackage{wrapfig}
\usepackage{booktabs}
\usepackage{multirow}

\usepackage[capitalize,noabbrev]{cleveref}

\usepackage{booktabs}   %用于三线表 \toprule, \midrule, \bottomrule
\usepackage{colortbl}   %用于单元格上色 \cellcolor
\usepackage{xcolor}     %用于颜色定义
\usepackage{multirow}   %如果你的表头有合并行
\usepackage{wrapfig}

\definecolor{bestclosed}{HTML}{E6CECF} 
\definecolor{bestopen}{HTML}{DAE8FC}
\definecolor{imprtext}{HTML}{C00000}

\hypersetup{hidelinks}

\title{Auxiliary Clues Aware's Geometry Problem Solving}

\author{%
  \textbf{Qigong Lei}$^{1,\ast}$ \quad
  \textbf{Qingsong Wang}$^{1,\ast}$ \quad
  \textbf{Wang Lin}$^{1}$ \quad
  \textbf{Qianyi Yang}$^{1}$\\
  \textbf{Chang Yao}$^{1,2}$ \quad
  \textbf{Jingyuan Chen}$^{1,\dagger}$\\
  \textit{$^1$Zhejiang University}\\
  \textit{$^2$Innovation and Management Center,}\\
  \textit{School of Software Technology (Ningbo), Zhejiang University}\\
  \texttt{\{leiqigong,wqsong,linwanglw,qianyiy\}@zju.edu.cn}\\
  \texttt{\{changy,jingyuanchen\}@zju.edu.cn}\\
  $^{\ast}$Equal contribution. \quad
  $^{\dagger}$Corresponding authors.
}

\begin{document}

\maketitle

\begin{abstract}
  Geometry Problem Solving (GPS) serves as a rigorous test of the complex reasoning capabilities of Multimodal Large Language Models (MLLMs), particularly when solutions require auxiliary constructions absent from the input diagram. However, explicit diagramming methods suffer from unstable image generation quality and cascading errors, while existing implicit attention mechanisms primarily focus on visible elements, limiting their ability to support reasoning that requires additional geometric constructions. To address these challenges, we propose \textbf{AuxCA} (Auxiliary-Clues-Aware GRPO), a reinforcement learning framework that helps models learn transferable auxiliary-construction strategies that can be applied beyond the training domain without explicit diagram generation at inference time. Specifically, we introduce Auxiliary Clues Dependency (\textbf{ACD}) to quantify the influence of auxiliary visual clues on token generation and integrate it into GRPO through fine-grained advantage weighting. Furthermore, we incorporate an auxiliary incentive reward to encourage the effective use of auxiliary geometric clues in correct solutions. Extensive experiments demonstrate that AuxCA-8B outperforms the explicit-plotting baseline Bagel-Canvas and achieves competitive performance with leading proprietary models on auxiliary-dependent plane geometry problems. Improvements on analytic geometry, solid geometry, and external benchmarks further suggest that the learned strategies can transfer beyond the training domain. Code is available at \url{https://github.com/fukua23/AuxCA}.
\end{abstract}

\begin{figure*}
  % \vspace{-5pt} % (可选) 微调垂直位置，有时需要向上移动一点
  \centering % 图片居中
  \includegraphics[width=1\textwidth]{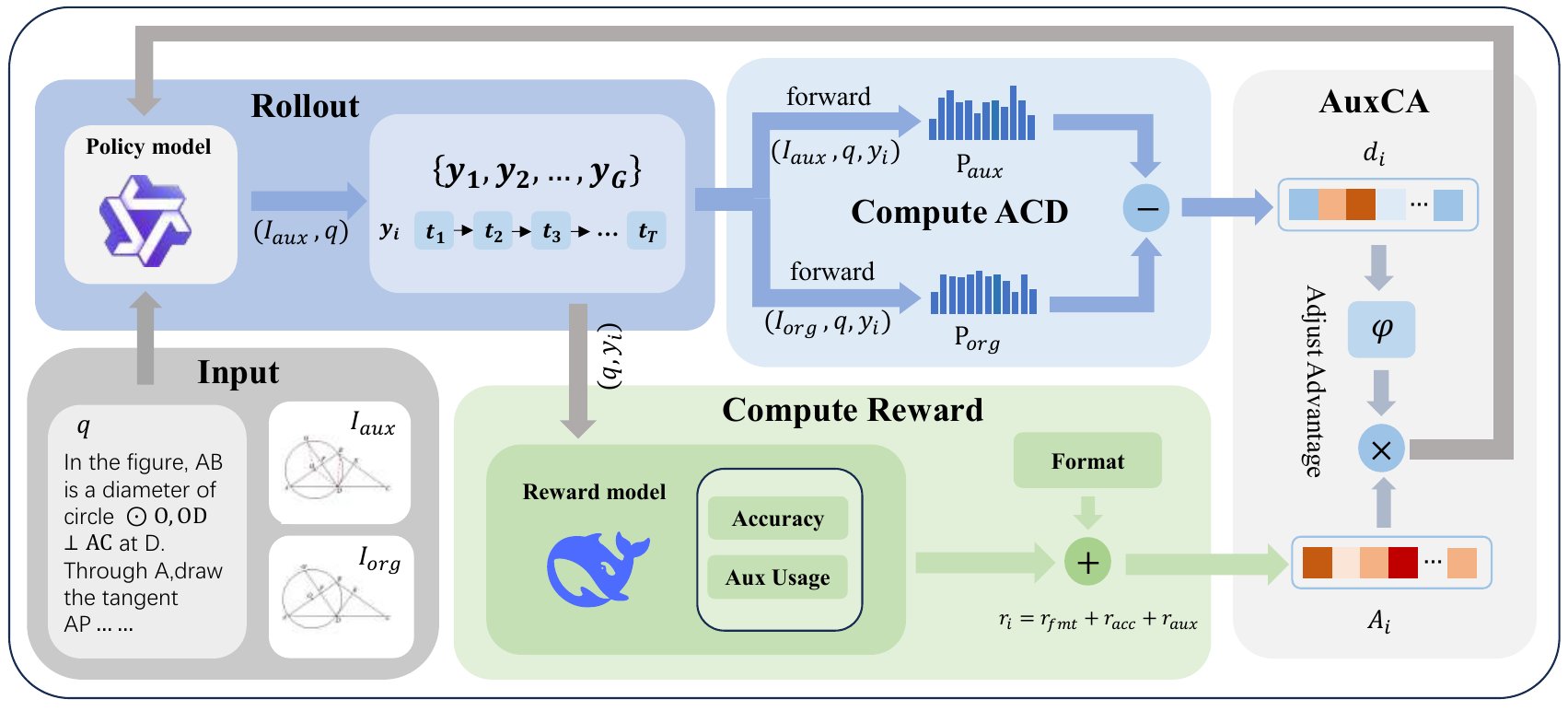}

\caption{
\textbf{The overall framework of AuxCA.} Following GRPO, the policy model first samples a group of outputs $\{y_1, \cdots , y_G\}$ conditioned on the auxiliary-enhanced input $I_{aux}$. We then calculate the token-level Auxiliary Clues Dependency $d_i$ by measuring the probability shift between predictions conditioned on $I_{aux}$ and those conditioned on $I_{org}$ without auxiliary visual clues. The sampled sequences are then scored by the reward model to derive the sequence-level advantages $A_i$. Finally, the model parameters are optimized by adjusting the advantage $A_i$ based on the dependency $d_i$ via the function $\psi$, thereby helping the model learn transferable auxiliary-construction strategies.}
\label{fig:main}
\vspace{-18pt}
\end{figure*}

\section{Introduction}

Geometry Problem Solving~\cite{ma2025survey, ping2025autogps, cho2026plane, xu2025geosense, huang2025autogeo} (GPS) is a key touchstone for evaluating MLLMs' complex reasoning capability, particularly for tasks requiring auxiliary construction. Driven by this challenge, recent works have transitioned from focusing on \textit{low-level visual perception} to \textit{mimicking high-level human cognition}. Early works, such as FM-ViT~\cite{lin2025non} and GeoX~\cite{xia2024geox}, primarily focused on optimizing visual encoders to precisely capture fine-grained geometric features like point-line relationships, or bridging the semantic gap through cross-modal alignment~\cite{zhuang2025math}. Subsequently, approaches represented by Visual Sketchpad~\cite{hu2024visual} and DeepEyes~\cite{zheng2025deepeyes} introduced the Visual Chain-of-Thought (VCoT), enabling models to mimic human behavior by explicitly generating or editing auxiliary diagrams to assist the reasoning process. Recently, inspired by the human cognitive mechanism of ``combining numbers and shapes'', MathCanvas~\cite{shi2025mathcanvas} proposed a two-stage training framework that further strengthens the model's VCoT capabilities in Mathematics. Trained under this framework, Bagel-Canvas generates intermediate diagrams through a diffusion-style rectified-flow process and achieved state-of-the-art performance on MathCanvas-Bench. By making geometric constructions explicit, this visual reasoning paradigm improves the interpretability of the reasoning process and supports more accurate solutions to complex geometry problems.

Despite the significant potential demonstrated by VCoT, this explicit visual generation paradigm faces severe challenges, including reliance on \textbf{external tool invocation}, \textbf{high inference costs}, \textbf{unstable image generation quality}, and \textbf{ error cascading}. To avoid the overhead of explicit diagramming, recent research, such as MINT-CoT~\cite{chen2025mint}, has begun to explore implicit visual enhancement paths, attempting to replace diagramming by improving the model's attention to specific visual patches. However, this approach suffers from a critical limitation in geometry problem-solving scenarios: the model can only attend to visual elements already present in the image. In contrast, the geometric solution process often requires introducing additional auxiliary clues (such as perpendicular lines or circumcircles) to derive the final answer. Furthermore, existing visual encoders lack prior perception of these ``potential geometric structures'', making it difficult for models to spontaneously focus on blank or implicit regions where auxiliary geometric structure should exist without guidance. In other words, relying solely on attention mechanisms is insufficient to enable the model to focus on the correct entry point for solving the problem from scratch.

To address these challenges, we propose \textbf{AuxCA (Auxiliary-Clues-Aware GRPO)}, a framework that helps models learn transferable auxiliary-construction strategies without generating auxiliary diagrams at inference time. Instead of relying on external plotting tools, AuxCA uses contrastive reasoning to identify \textbf{regions containing key auxiliary clues}. Specifically, we introduce \textbf{Auxiliary Clues Dependency (ACD)}, which measures the probability shift in generating the same answer when the model is conditioned on images with auxiliary clues versus the original images. In this way, ACD quantifies the dependency of each generated token on auxiliary visual information. Tokens whose probabilities increase significantly in the presence of auxiliary clues are regarded as closely associated with the additional geometric structures required for solving the problem. Based on ACD, AuxCA applies \textbf{fine-grained advantage weighting} to high-dependency tokens during GRPO-based reinforcement learning~\cite{shao2024deepseekmath}. By emphasizing tokens that depend on auxiliary clues, this mechanism helps the model learn auxiliary-construction strategies that can generalize to unseen geometry problems. Moreover, we assign an additional reward when the model answers correctly while leveraging such geometric information, encouraging it to actively attend to latent auxiliary clues during reasoning. As a result, AuxCA achieves higher reasoning efficiency than explicit Visual CoT without requiring explicit diagramming.

The main contributions of this paper are as follows:

\begin{itemize}
    \item We introduce Auxiliary Clues Dependency (ACD), a token-level measure of how auxiliary geometric clues change the probability of a generated token.
    \item Based on ACD, we propose AuxCA, a framework that reweights token-level advantages in GRPO. This helps the model learn auxiliary-construction strategies that can generalize to unseen geometry problems without relying on explicit plotting at inference time.
    \item We train AuxCA-8B and demonstrate substantial gains on auxiliary-dependent plane geometry, with improvements extending to analytic geometry, solid geometry, and external benchmarks.
\end{itemize}

\section{Related Work}

\textbf{Geometry Problem Solving.} 
Research in the GPS~\cite{trinh2024solving, cho2025geodano, zhou2025solving, lin2025goedel, lin2025goedel_1, zhang2024mavis} has witnessed significant progress in recent years, with its primary evolutionary trajectory shifting from enhancing low-level visual perception and optimizing cross-modal alignment toward unleashing models' active visual reasoning and generative capabilities. Addressing the limitations of traditional visual encoders in capturing high-frequency details of geometric figures, such as point-line relationships, FM-ViT~\cite{lin2025non} and GeoX~\citep{xia2024geox} have significantly improved perception quality through frequency-domain modulation mechanisms and domain-specific masked autoencoder pre-training, respectively. Meanwhile, Math-PUMA~\cite{zhuang2025math} leverages the strong logical capabilities of plain text to supervise the visual modality, narrowing the semantic gap between modalities via progressive alignment. To mimic the human cognitive process of ``thinking while drawing'', Visual Sketchpad~\cite{hu2024visual} and MathCanvas~\cite{shi2025mathcanvas} introduced the Visual Chain-of-Thought (VCoT), enabling models to explicitly facilitate reasoning by generating or editing auxiliary lines and diagrams. Furthermore, MINT-CoT~\cite{chen2025mint} overcomes the limitations of coarse-grained bounding boxes through an interleaved visual token selection mechanism, achieving pixel-level fine-grained interaction between image and text. Finally, inspired by DeepSeek-R1~\cite{shao2024deepseekmath}, works such as DeepEyes~\cite{zheng2025deepeyes}, LMM-R1~\cite{peng2025lmm}, and VisTA~\cite{huang2025visualtoolagent} have incorporated reinforcement learning into multimodal reasoning. By utilizing outcome feedback to incentivize active visual exploration, self-correction, and autonomous tool usage, these approaches are propelling the field from mere perceptual understanding toward verifiable, complex logical reasoning.

\textbf{Improvements to GRPO.} 
% Although GRPO significantly lowers the resource barrier of reinforcement learning by discarding the critic model, its limitations regarding optimization stability, sample bias, and fine-grained credit assignment have prompted a wave of improvements.
To address the issues of restricted gradient updates and low sample efficiency in large-scale training, DAPO~\cite{yu2025dapo} proposed a Decoupled Clip mechanism to allow for larger policy updates on advantaged samples. Combined with a Dynamic Sampling strategy, this approach effectively accelerates training convergence. Meanwhile, to tackle the ``difficulty bias'', where simple samples dominate gradient updates during the later stages of training, UniVG-R1~\cite{bai2025univg} introduced a Difficulty-Aware Weight Adjustment strategy. By dynamically scaling gradients based on sample performance, it successfully incentivizes models to conquer more challenging samples in universal visual grounding tasks. Furthermore, in terms of enhancing the precision of complex reasoning and multimodal alignment, researchers have identified the drawback of GRPO applying uniform optimization pressure across all tokens. HICRA~\cite{hicra} reveals the hierarchical distinction between high-level strategic planning and low-level program execution, resolving the issue of learning signal dilution via hierarchy-aware credit assignment. VICRA~\cite{hong2026dont} addresses the bottleneck of stagnating visual perception capabilities in MLLMs; by quantifying the visual dependency of tokens and reallocating optimization weights, it effectively corrects the optimization imbalance between textual and visual modalities. In addition, TR-GRPO~\cite{le2025tokenregulated} places greater emphasis on high-probability tokens, whereas TW-GRPO~\cite{dang2025reinforcingvideoreasoningfocused} improves GRPO by assigning higher rewards to high-entropy tokens.

\section{Method}
\label{sec:method}
In this section, we present the details of our proposed method, AuxCA. In Section~\ref{sec:ACD}, we define the auxiliary clues dependency for tokens. Section~\ref{sec:aea_grpo} describes how we utilize this dependency to perform fine-grained optimization on GRPO, thereby enhancing the model's capability in solving geometry problems. Finally, Section~\ref{sec:reward} details the reward design tailored for GPS scenarios.
\subsection{Auxiliary Clues Dependency}
\label{sec:ACD}
The geometry problem-solving task is typically formalized as follows: given the problem text $q$ and the original image $I_{\text{org}}$, the goal is for the model $M$ to generate the corresponding solution $y$, denoted as $y=M(I, q)$. However, when dealing with complex geometry problems, introducing an image containing auxiliary clues, denoted as $I_{\text{aux}}$, often plays a decisive role. It provides critical directional guidance and significantly improves solution accuracy. This improvement essentially stems from the model focusing more on specific tokens that carry important auxiliary information during generation. To capture this characteristic, we define Auxiliary Clues Dependency (ACD) to measure the sensitivity of each token to auxiliary geometric elements during the generation process.

We construct ACD by leveraging the probability shift of specific tokens generated by the model under different visual inputs ($I_{\text{org}}$ versus $I_{\text{aux}}$). First, we input the image with auxiliary clues $I_{\text{aux}}$ and the problem text $q$ into the model to obtain the target solution sequence $y_{\text{aux}}$ under auxiliary information enhancement, along with the probability of each token $\pi_\theta(y_{\text{aux},t} \mid I_{\text{aux}}, q, y_{\text{aux},<t})$. Subsequently, to deconstruct the contribution of auxiliary clues, we perform counterfactual inference. We measure the likelihood of the model generating the same token $y_{\text{aux},t}$ without auxiliary hints (\textit{i.e.}, using the original image $I_{\text{org}}$). To do this, while keeping the solution sequence $y_{\text{aux}}$ unchanged, we replace the input image with $I_{\text{org}}$ and perform forward propagation again to calculate the corresponding conditional probability $\pi_\theta(y_{\text{aux},t} \mid I_{\text{org}}, q, y_{\text{aux},<t})$. Finally, we quantify the dependency of the token generated at step $t$ on auxiliary clues using Eq.~\ref{eq:d}:

\begin{equation}
\label{eq:d}
\begin{split}
d_{t} = \pi_\theta(y_{\text{aux},t} \mid I_{\text{aux}}, q, y_{\text{aux},<t})- 
\pi_\theta(y_{\text{aux},t} \mid I_{\text{org}}, q, y_{\text{aux},<t}),
\end{split}
\end{equation}

where $d_t \in [-1, 1]$. A value closer to 1 indicates that the generation of this token relies heavily on the extra visual clues provided by the auxiliary image $I_{\text{aux}}$. Conversely, a smaller $d_{t}$ suggests that the generation of the token does not depend on auxiliary clues. In the next section, we will detail how to utilize ACD to optimize the problem-solving capability of the model.

\subsection{Auxiliary-Clues-Aware GRPO}
\label{sec:aea_grpo}
The success of DeepSeek-Math~\cite{shao2024deepseekmath} validates the efficiency of GRPO. Formally, for a given input, GRPO samples a group of candidate answers $\{y_1, y_2, \cdots, y_G\}$ from the policy model, where $G$ denotes the group size. Subsequently, each generated answer is evaluated by a rule-based validator or a reward model to obtain rewards $\{r_1, r_2, \cdots, r_G\}$. The advantage value $A_i$ for each answer is then computed by normalizing the rewards against the group reward mean. The training objective of GRPO is to maximize the expectation of these relative advantages, as formulated below:

\begin{equation}
\label{eq:GRPO}
\begin{split}
\mathcal{J}_{\text{GRPO}}(\theta) &= \mathbb{E}_{(I, q) \sim P(Q), \{y_i\} \sim \pi_{\theta_{\text{old}}}} \Bigg[
\quad \frac{1}{G} \sum_{i=1}^G \frac{1}{|y_i|} \sum_{t=1}^{|y_i|} \Bigg\{ \\
&\min \Bigg( \frac{\pi_\theta(y_{i,t} | \cdot)}{\pi_{\theta_{\text{old}}}(y_{i,t} | \cdot)} A_{i,t}, \quad \operatorname{clip}\left(\frac{\pi_\theta(y_{i,t} | \cdot)}{\pi_{\theta_{\text{old}}}(y_{i,t} | \cdot)},1-\epsilon,1+\epsilon \right) A_{i,t} \Bigg) - \beta \, \mathbb{D}_{\text{KL}}[\pi_\theta \parallel \pi_{\text{ref}}] \Bigg\} \Bigg],
\end{split}
\end{equation}

where $\varepsilon$ and $\beta$ are hyperparameters. $A_{i,t}$ represents the relative advantage, calculated as:

\begin{equation}
A_{i,t} = \frac{r_i - \mathrm{mean}(\{r_1, r_2, \cdots, r_G\})}{\mathrm{std}(\{r_1, r_2, \cdots, r_G\})},
\end{equation}

As seen in Eq.~\ref{eq:GRPO}, the original GRPO implicitly assumes that all tokens within the same answer sequence contribute equally, thus assigning them identical advantage values. However, in geometry reasoning tasks, the importance of tokens is distributed unevenly. Tokens that carry not only semantic information but also critical clues about auxiliary geometric constructions often play a decisive role in the solution path.

To address this limitation, we propose a Token-level Fine-grained Advantage Weighting strategy, leveraging the ACD. The core idea is to dynamically adjust the advantage value of a token based on its dependency on auxiliary clues during generation. Specifically, if a token relies heavily on auxiliary information (indicated by a high ACD), we amplify its advantage value. Emphasizing these key tokens helps the model learn auxiliary-construction strategies that can generalize to unseen geometry problems without requiring explicit diagram generation at inference time. Conversely, for tokens with low dependency, the advantage is suppressed. This mechanism aims to guide the model to utilize auxiliary clues more effectively.

The specific implementation process is illustrated in Figure~\ref{fig:main}. First, we sample a group of responses from the policy model $M$ given the input  $(q, I_{\text{aux}}, I_{\text{org}})$, following the original GRPO formulation. Subsequently, we calculate the auxiliary clues dependency $\{\textbf{d}_1, \textbf{d}_2, \cdots, \textbf{d}_G\}$ for each generated sequence. Next, we introduce a mapping function  $\psi$ to transform the dependency obtained in the previous step into a relative advantage weighting coefficient $\psi_{i,t}$ for each token. Finally, we adjust the original advantage values token-by-token using this coefficient. The corrected advantage value is:

\begin{equation}
A_{i,t}^{\mathrm{AuxCA}} = \psi(d_{i,t})\cdot A_{i,t},
\end{equation}

where $\psi$ is defined as:

\begin{equation}
\psi(d_{i,t}) = 1 + \left( d_{i,t} - \frac{1}{|y_i|} \sum_{t=1}^{|y_i|} d_{i,t} \right).
\end{equation}

Through mathematical derivation (refer to the Appendix~\ref{sec:proof_conservation} for details), we demonstrate that this weighting strategy possesses the property of Total Advantage Conservation. That is, for the same answer sequence, the cumulative advantage value remains unchanged before and after reweighting:

\begin{equation}
\label{eq:consitant}
\sum_{t=1}^{|y_i|}A_{i,t} = \sum_{t=1}^{|y_i|} \psi(d_{i,t})\cdot A_{i,t}.
\end{equation}

This equation indicates that the function of $\psi$ is not to scale the global advantage signal of the entire answer. Instead, it performs a fine-grained redistribution of advantage within the sequence while keeping the total amount constant. Through comparison with another exponential function (see Appendix~\ref{sec:appendix_psi_ablation}), we find that this property can effectively prevent training instability caused by advantage drift, which is the core motivation for selecting this form as the mapping function.

\subsection{Reward Design}
\label{sec:reward}

Due to the unique nature of geometry proof problems, traditional rule-based matching reward functions are often inapplicable. On one hand, questions typically contain multiple sub-problems; on the other hand, the correct solution path is often not unique. To address this, we design a multi-dimensional reward mechanism that covers correctness, auxiliary clues incentives, and format specifications.

First, given the complexity and openness of the solution space, we abandon simple rule matching in favor of using an LLM as a judge. Following the standards of MathCanvas-Bench, we score each sub-problem independently. If the model answers all sub-problems correctly, it receives the full reward; if only a subset is correct, it receives partial points based on weights. This dense feedback signal enables the model to obtain positive reinforcement from partially correct reasoning, even if it has not fully solved the entire problem. Detailed prompts are provided in the Appendix~\ref{sec:rewardprompt}.

Second, the ACD-based advantage weighting strategy in Section~\ref{sec:aea_grpo} helps the model learn transferable auxiliary-construction strategies, but it does not explicitly reward the effective use of these clues in a correct solution. We need to encourage the model to leverage these geometric clues for deduction, thereby achieving high accuracy without relying on explicit diagramming. To this end, we introduce a sparse reward term $r_{\text{aux}}$. Specifically, we employ an LLM to examine the reasoning text generated by the model. This reward is triggered only when the model satisfies two conditions simultaneously: answering all sub-problems correctly and explicitly referencing the geometric relationships provided by the auxiliary clues. This design reinforces the value of auxiliary clues at a macro behavioral level. Experiments (see Appendix~\ref{sec:appendix_hyperparam_3}) indicate that model performance is optimal when $r_{\text{aux}} = 0.2$.

Finally, to enforce output structure, we introduce $r_{\text{fmt}}$ to guide the model to adhere to predefined output specifications (such as using \texttt{<think>} tags). Consequently, the final reward function is defined as:

% \vspace{-15pt}
\begin{equation}
r = r_{\text{fmt}} + r_{\text{acc}} + r_{\text{aux}}.
\end{equation}
% \vspace{-10pt}

\begin{table*}[]

\caption{\textbf{Main results on MathCanvas-Bench.} We report the Complete and Weighted accuracy scores across three subsets (Aux-PlaGeo, AnaGeo, and SolidGeo) and the overall dataset. ``\ding{51}'' in the ``Think'' column indicates models that utilize chain-of-thought or reasoning tokens. The best \colorbox{bestclosed}{closed-source} and \colorbox{bestopen}{open-source} highest accuracy of MLLMs are marked in red and blue, respectively. 
\label{table:main}}

\centering
\resizebox{\linewidth}{!}{
% 定义列格式：左对齐 | 居中 | 居中 | pg_subset(2列) | ana_geo(2列) | solid_geo(2列) | overall(2列)
\begin{tabular}{lcccccccccc}
\toprule
\multirow{2}{*}{\textbf{Model}} & \multirow{2}{*}{\textbf{Size}} & \multirow{2}{*}{\textbf{Think}} & \multicolumn{2}{c}{\textbf{Aux-PlaGeo}} & \multicolumn{2}{c}{\textbf{AnaGeo}} & \multicolumn{2}{c}{\textbf{SolidGeo}} & \multicolumn{2}{c}{\textbf{overall}} \\ 
\cmidrule(lr){4-5}\cmidrule(lr){6-7}\cmidrule(lr){8-9}\cmidrule(lr){10-11}
 &  &  & Complete & Weighted & Complete & Weighted & Complete & Weighted & Complete & Weighted \\ \hline
\multicolumn{11}{c}{\small{\textit{Closed Source}}} \\ \hline
Gemini-2.5-Pro & - & \ding{51} & \cellcolor{bestclosed}40.2 & \cellcolor{bestclosed}54.5 & 47.4 & \cellcolor{bestclosed}58.6 & 35.4 & \cellcolor{bestclosed}48.8 & \cellcolor{bestclosed}48.3 & \cellcolor{bestclosed}59.4 \\
Gemini-2.5-Flash & - & \ding{55} & 29.4 & 39.8 & 44.9 & 55.7 & 31.2 & 40.6 & 38.9 & 48.8 \\
GPT-4.1 & - & \ding{55} & 7.7 & 18.7 & 20.9 & 30.3 & 8.9 & 23.8 & 18.2 & 28.9 \\
GPT-4o & - & \ding{55} & 7.1 & 17.1 & 10.5 & 18.2 & 5.8 & 18.1 & 9.8 & 18.8 \\
GPT-5 & - & \ding{51} & 32.4 & 43.8 & \cellcolor{bestclosed}47.8 & 56.2 & 26.1 & 35.4 & 42.7 & 50.9 \\
Claude-Sonnet-4 & - & \ding{51} & 24.1 & 31.8 & 29.4 & 37.2 & 22.8 & 32.4 & 24.3 & 36.4 \\
Seed-1.6-Thinking & - & \ding{51} & 39.8 & 53.3 & 42.3 & 57.6 & \cellcolor{bestclosed}36.1 & 45.9 & 44.5 & 55.3 \\
Qwen3-VL-Plus & - & \ding{51} & 34.6 & 44.5 & 41.6 & 54.3 & 34.7 & 43.2 & 41.2 & 52.1 \\ \hline
\multicolumn{11}{c}{\small{\textit{Open Source}}} \\ \hline
Qwen-2.5-VL-Instruct & 7B & \ding{55} & 7.2 & 16.9 & 9.7 & 16.4 & 6.0 & 19.1 & 9.2 & 17.9 \\
Qwen-3-VL-Instruct & 8B & \ding{55} & 13.1 & 23.0 & 24.4 & 34.9 & 11.3 & 25.6 & 19.6 & 29.6 \\
Qwen-3-VL-Thinking & 8B & \ding{51} & 18.4 & 24.6 & 28.1 & 33.4 & 14.6 & 27.4 & 26.9 & 33.3 \\
Qwen-3-VL-Thinking & 32B & \ding{51} & 28.3 & 35.1 & \cellcolor{bestopen}43.3 & \cellcolor{bestopen}49.2 & \cellcolor{bestopen}23.9 & 36.3 & \cellcolor{bestopen}36.9 & \cellcolor{bestopen}44.4 \\
Gemma-3-it & 27B & \ding{55} & 11.3 & 22.5 & 16.8 & 25.8 & 7.6 & 21.4 & 14.9 & 25.4 \\
InternVL3.5 & 8B & \ding{51} & 14.5 & 19.3 & 22.8 & 27.8 & 11.7 & 22.8 & 20.9 & 26.9 \\
InternVL3.5-A3B & 30B & \ding{51} & 19.8 & 30.4 & 36.1 & 43.8 & 16.1 & 33.0 & 28.8 & 39.3 \\
Keye-VL-1.5 & 8B & \ding{51} & 18.2 & 27.1 & 15.8 & 31.4 & 13.4 & 27.9 & 22.2 & 31.2 \\
Bagel-Canvas & 14B & \ding{55} & 27.2 & 39.9 & 16.0 & 27.2 & 18.0 & 35.3 & 21.9 & 34.4 \\\hline
\textbf{AuxCA} & 8B & \ding{51} & \cellcolor{bestopen}31.7 & \cellcolor{bestopen}44.6 & 31.2 & 41.0 & 22.8 & \cellcolor{bestopen}40.4 & 30.3 & 42.2 \\ 
$\Delta$ Over Base Model & $ $ & $ $ & \textcolor{imprtext}{+13.3} & \textcolor{imprtext}{+20.0} & \textcolor{imprtext}{+3.1} & \textcolor{imprtext}{+7.6} & \textcolor{imprtext}{+8.2} & \textcolor{imprtext}{+13.0} & \textcolor{imprtext}{+3.4} & \textcolor{imprtext}{+8.9} \\ \hline
\end{tabular}
}
\end{table*}

\section{Experiments}

%\subsection{Experimental setup}

\textbf{Dataset}
This study focuses on solving plane geometry problems that are \textbf{Auxiliary-dependent}. To this end, we filtered and constructed a dedicated training set containing 70k samples from the plane geometry subset of MathCanvas-Instruct~\cite{shi2025mathcanvas}. All these samples require the introduction of additional auxiliary geometric structures to complete the reasoning process (detailed selection are provided in the Appendix~\ref{sec:appendix_data_construction} and Appendix~\ref{sec:data_pipeline}).

\textbf{Benchmark}
During evaluation, we extracted 635 data points requiring the addition of extra auxiliary elements to the original image from MathCanvas-Bench to serve as the core test set (\texttt{Aux-PlaGeo}). Furthermore, to comprehensively measure model performance, we not only conducted extensive evaluations on two other related subsets Analytical geometry (\texttt{AnaGeo}), Solid geometry (\texttt{SolidGeo}) and the full set (\texttt{overall}) of this benchmark (see Table~\ref{table:main}), but also performed generalization analysis on mainstream benchmarks such as MathVista~\cite{lu2024mathvista}, MathVerse~\cite{zhang2024mathverse}, and MathVision~\cite{wang2024measuring,wang2025mathcodervl} (see Table~\ref{tab:generalization_final_fix}). Performance is measured via Complete Accuracy, reflecting a strict all-or-nothing criterion for complete solutions, and Weighted Accuracy, which captures partial progress through fine-grained exponential weights.

\textbf{Baseline}
 To provide a comprehensive evaluation of AuxCA, we benchmarked it against leading state-of-the-art proprietary models, such as the Gemini-2.5~\cite{comanici2025gemini} series (Gemini-2.5-Pro, Gemini-2.5-Flash), the ChatGPT series (ChatGPT-4o~\cite{hurst2024gpt}, ChatGPT-4.1~\cite{hurst2024gpt}, ChatGPT-5~\cite{singh2025openai}), Claude-Sonnet-4\footnote{\url{https://www.anthropic.com/claude/sonnet}}, Seed-1.6-Thinking\footnote{\url{https://seed.bytedance.com/en/seed1_6}}, and Qwen3-VL-Plus~\cite{bai2025qwen3vltechnicalreport}. Additionally, we included a diverse range of mainstream open-source models with parameters spanning from 7B to 32B. This selection specifically covers the Qwen series (Qwen-2.5-VL-Instruct~\cite{bai2025qwen25vltechnicalreport}, Qwen-3-VL-Instruct/Thinking~\cite{bai2025qwen3vltechnicalreport}), the InternVL series (InternVL3.5-8B/30B)~\cite{wang2025internvl3}, Gemma-3-it~\cite{team2025gemma}, Keye-VL-1.5~\cite{yang2025kwai}, and Bagel-Canvas~\cite{shi2025mathcanvas}.

\textbf{Implementation Details}
The base model used in the experiments is Qwen3-VL-8B-Thinking, while the reward model is DeepSeek-V3.2. All experiments were implemented using the \texttt{trl}~\cite{vonwerra2020trl} framework; specific hyperparameter settings and additional implementation details are provided in Appendix~\ref{sec:appendix_implementation}. To ensure training stability, we performed Supervised Fine-Tuning (SFT) on the model for 2 epochs using the aforementioned 70k dataset. Subsequently, we randomly sampled 15k samples from this dataset for AuxCA reinforcement learning. Through this training, the model improves its sensitivity to auxiliary geometric elements and learns transferable auxiliary-construction strategies that can be applied beyond the training domain. We name the resulting model \textbf{AuxCA-8B}.

\subsection{Main Experiments}
Table~\ref{table:main} presents the performance of our model, AuxCA-8B, across various subsets and the full MathCanvas-Bench. The experimental results provide strong validation for the effectiveness of the reinforcement learning strategy proposed in this paper.

% \begin{figure}
%   % \vspace{-5pt} % (可选) 微调垂直位置，有时需要向上移动一点
%   \centering % 图片居中
%   \includegraphics[width=1\columnwidth]{img/bar_metrics.pdf}

% \caption{
% \textbf{Analysis of auxiliary clues internalization.} \textit{GRPO($I_{\mathrm{aux}}$)} is evaluated with images containing correct auxiliary clues at test time, while all other methods use the original problem images. \textit{aux acc} denotes the fraction of samples for which the auxiliary clues are correctly constructed.
% }
% \label{fig:hitmap}
% % \vspace{-10pt}
% \end{figure}

\begin{wrapfigure}{r}{0.48\linewidth}
    \vspace{-10pt}
    \centering
    \includegraphics[width=\linewidth]{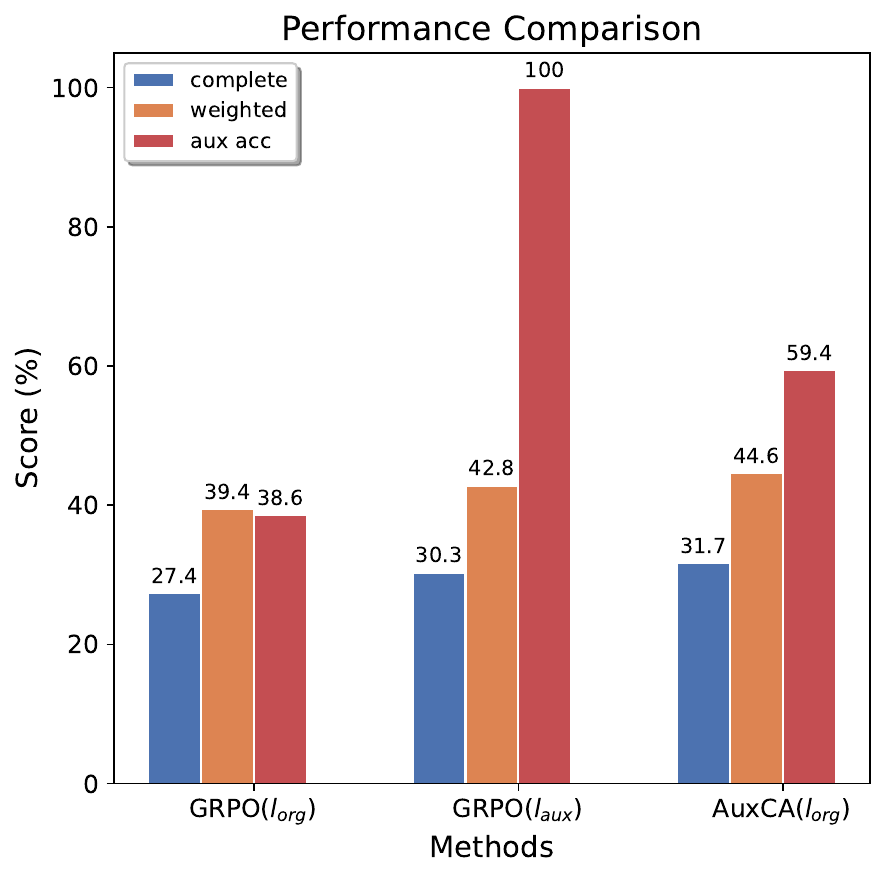}
    \vspace{-16pt}
    \caption{
    \textbf{Analysis of auxiliary construction and utilization.} \textit{GRPO($I_{\mathrm{aux}}$)} is evaluated with images containing correct auxiliary clues at test time, while all other methods use the original problem images. \textit{aux acc} denotes the fraction of samples for which the auxiliary clues are correctly constructed.
    }
    \label{fig:bar_metrics}
    \vspace{-10pt}
\end{wrapfigure}

\textbf{Learning Transferable Auxiliary-Construction Strategies}
As shown in Table~\ref{table:main}, AuxCA consistently outperforms other models with comparable parameter sizes across all benchmarks, as well as larger models such as Bagel-Canvas, InternVL3.5-30B-A3B, and Gemma-3-27b-it. On the \texttt{Aux-PlaGeo} dataset, AuxCA achieves a substantial improvement of +20.0 in the Weighted Score (from 24.6 to 44.6). Beyond its strong performance in the training domain of plane geometry, AuxCA also demonstrates generalization to analytic geometry (\texttt{AnaGeo}) and solid geometry (\texttt{SolidGeo}). Specifically, on \texttt{SolidGeo}, AuxCA improves the Weighted Score by +13.0 over the base model, reaching 40.4. This result surpasses that of Qwen-3-VL-Thinking 32B (36.3), a model with four times more parameters. These performance gains suggest that incorporating the Auxiliary Clues Dependency (ACD) mechanism and a fine-grained advantage reweighting strategy helps AuxCA learn transferable auxiliary-construction strategies that can be applied beyond the training domain.

To further assess the model's ability to propose and use auxiliary constructions, we evaluate the GRPO-trained model with an image augmented with correct auxiliary clues, denoted as $I_{\text{aux}}$, in place of the original input image $I_{\text{org}}$. We use this setting as a strong baseline. We also measure the proportion of samples in which AuxCA and GRPO correctly construct auxiliary clues when reasoning from the original image, referred to as auxiliary construction accuracy (aux acc). The results are shown in Figure~\ref{fig:bar_metrics}. After AuxCA-based reinforcement learning, the model achieves a 20.8\% higher auxiliary construction accuracy than GRPO, suggesting that AuxCA helps the model learn more effective auxiliary-construction strategies. Moreover, even when provided with correct auxiliary clues, GRPO still performs slightly worse than AuxCA, which uses only the original image. Together, these results suggest that AuxCA training improves the model's ability to propose useful auxiliary constructions and reason with the resulting geometric relationships.

\textbf{Superiority over Explicit Plotting}
To compare auxiliary-construction strategies expressed through textual reasoning with explicit plotting, we evaluate AuxCA-8B against Bagel-Canvas (14B with 7B active parameters), a strong explicit-plotting baseline. Bagel-Canvas generates auxiliary diagrams to assist problem-solving, whereas AuxCA expresses auxiliary constructions through chain-of-thought reasoning without generating additional images. Despite having fewer total parameters (8B vs. 14B), AuxCA outperforms Bagel-Canvas on both Complete and Weighted Scores across all test subsets of MathCanvas-Bench. Notably, on the Weighted Score of the full set (\texttt{overall}), AuxCA achieves 42.2, compared with 34.4 for Bagel-Canvas. Under the same hardware and inference settings, AuxCA also reduces average end-to-end latency by 65.4\%, despite generating more text tokens; details are provided in Appendix~\ref{sec:inference_cost}. These results suggest that learning auxiliary-construction strategies through textual reasoning offers an effective alternative to explicit plotting, achieving higher solution accuracy in these evaluations while avoiding the need for additional image generation.

\textbf{Comparison with Leading Closed-Source Models}
On the \texttt{Aux-PlaGeo} subset, which specifically evaluates problems requiring auxiliary constructions, AuxCA-8B achieves performance competitive with leading closed-source models. Its Weighted Score (44.6) exceeds those of GPT-4o (17.1), Claude-Sonnet-4 (31.8), and GPT-5 (43.8). These results suggest that domain-specific training data and targeted reinforcement learning can help smaller models learn effective auxiliary-construction strategies and achieve competitive performance on challenging geometry reasoning tasks. Together with the gains on \texttt{AnaGeo} and \texttt{SolidGeo}, they provide evidence that these strategies can transfer beyond the plane geometry training domain.

\vspace{-3pt}

\begin{table}[t]
    \centering
    \caption{\textbf{Generalization capabilities on external benchmarks.}}
    \label{tab:generalization_final_fix}

    \small
    \renewcommand{\arraystretch}{1.25}

    \begin{minipage}[t]{0.30\linewidth}
        \centering
        \vspace{2pt}
        \textbf{(a) MathVista}\\[2pt]
        \setlength{\tabcolsep}{6pt}
        \begin{tabular}{lc}
            \toprule
            Model & GPS \\
            \midrule
            Base & 68.7 \\
            \textbf{AuxCA} & \textbf{71.6} \\
            Improve & \textcolor{red}{+2.9} \\
            \bottomrule
        \end{tabular}
    \end{minipage}
    \hfill
    \begin{minipage}[t]{0.33\linewidth}
        \centering
        \vspace{2pt}
        \textbf{(b) MathVerse}\\[2pt]
        \setlength{\tabcolsep}{4pt}
        \begin{tabular}{lcc}
            \toprule
            Model & Text Dom & Text Lite \\
            \midrule
            Base & 69.3 & 65.1 \\
            \textbf{AuxCA} & \textbf{76.5} & \textbf{67.0} \\
            Improve & \textcolor{red}{+7.2} & \textcolor{red}{+1.9} \\
            \bottomrule
        \end{tabular}
    \end{minipage}
    \hfill
    \begin{minipage}[t]{0.34\linewidth}
        \centering
        \vspace{2pt}
        \textbf{(c) MathVision}\\[2pt]
        \setlength{\tabcolsep}{4pt}
        \begin{tabular}{lccc}
            \toprule
            Model & Angle & Area & Length \\
            \midrule
            Base & 22.5 & 21.2 & 23.2 \\
            \textbf{AuxCA} & \textbf{31.0} & \textbf{25.8} & \textbf{28.9} \\
            Improve & \textcolor{red}{+8.5} & \textcolor{red}{+4.6} & \textcolor{red}{+5.7} \\
            \bottomrule
        \end{tabular}
    \end{minipage}
    \vspace{-10pt}
\end{table}

\begin{wraptable}{r}{0.48\linewidth}
\vspace{-10pt}
\centering
\caption{\textbf{Ablation study on the core components of AuxCA.}}
\label{table:ablation}

\vspace{-1pt}
\small
\renewcommand{\arraystretch}{1.15}
\setlength{\tabcolsep}{5pt}

\begin{tabular}{lcc}
\toprule
\multirow{2}{*}{Model} & \multicolumn{2}{c}{Aux-PlaGeo} \\
\cmidrule(lr){2-3}
 & Complete & Weighted \\
\midrule
AuxCA              & \textbf{31.7} & \textbf{44.6} \\
- w/o ACD(GRPO)  & 27.4          & 39.4          \\
- w/o $r_{aux}$  & 25.8           & 38.1          \\
- SFT only & 22.1          & 33.9          \\
Base model          & 18.4          & 24.6          \\
\bottomrule
\end{tabular}
\vspace{-10pt}
\end{wraptable}

\subsection{Ablation Study}

To validate the independent contributions of each core component in our framework, we conduct ablation studies as shown in Table~\ref{table:ablation}. First, we examine the effect of removing the Auxiliary Clues Dependency (ACD) mechanism (w/o ACD). Without this module, the algorithm degenerates into the standard GRPO with uniform advantage assignment, resulting in a drop of 5.2 in the Weighted Score (from 44.6 to 39.4) and a decrease of 4.3 in the Complete Score (from 31.7 to 27.4). These results support the effectiveness of our fine-grained reweighting strategy in improving geometric reasoning. Complementary analysis in Appendix~\ref{sec:acd_scores} shows that 97 of the 100 highest-scoring tokens are manually identified as auxiliary-clue or construction-action tokens, providing further evidence that ACD captures content relevant to auxiliary construction. Next, we evaluate the impact of removing the auxiliary reward term (w/o $r_{aux}$). The complete score drops from 31.7 to 25.8, and the weighted score decreases from 44.6 to 38.1, indicating that an explicit reward signal is crucial for guiding the model to actively utilize auxiliary information during reasoning. Finally, we compare AuxCA with the SFT-only baseline without subsequent reinforcement learning, which achieves lower scores on both metrics, further supporting the effectiveness of the AuxCA reinforcement learning stage.

We attribute this to the fact that SFT and GRPO are trained directly with $I_{\text{aux}}$ as input, in which the correct auxiliary clues are already provided in the image. As a result, these models primarily learn to reason using the given auxiliary information. In contrast, AuxCA leverages both $I_{\text{org}}$ and $I_{\text{aux}}$ to adjust token-level advantages. This not only strengthens the model’s ability to reason effectively with correct auxiliary information, but also enhances its capability to construct auxiliary clues.
\vspace{-5pt}

% \subsection{Generalization Analysis}
% To further verify the generalization capability of AuxCA, we conducted zero-shot testing on MathVista, MathVerse, and MathVision (as shown in Table~\ref{tab:generalization_final_fix} ). The experimental results demonstrate that AuxCA significantly outperforms the base model across all benchmarks. Specifically, on the Text Dominant subset of MathVerse, the model achieved a substantial improvement of +7.2, which is considerably higher than the gain on the Text Lite subset. This suggests that the auxiliary-construction strategies learned during training may also facilitate reasoning in settings with limited visual information. Furthermore, fine-grained analysis on MathVision reveals that the model made the most evident progress in tasks heavily dependent on auxiliary construction, such as Angle (+8.5), Length (+5.7), and Area (+4.6). This strongly corroborates that AuxCA has transformed specific auxiliary construction strategies (\textit{e.g.}, drawing altitudes, extending segments) into general geometric problem-solving intuition (Geometric Intuition). This enables the model to maintain exceptional robustness and reasoning precision when facing diverse geometric challenges. We also evaluate the performance of AuxCA-8B on problems that do not require additional auxiliary clue construction, as shown in the Appendix~\ref{sec:no_aux}.

\subsection{Generalization Analysis}

To further evaluate the generalization capability of AuxCA, we conduct zero-shot evaluations on MathVista, MathVerse, and MathVision, as shown in Table~\ref{tab:generalization_final_fix}. The results demonstrate that AuxCA consistently outperforms the base model across all three benchmarks, indicating that the learned auxiliary-construction strategies can transfer to unseen problems. Specifically, on the Text Dominant subset of MathVerse, AuxCA achieves a 7.2-point improvement, substantially exceeding the gain observed on the Text Lite subset. This suggests that the auxiliary-construction strategies learned during training may also facilitate reasoning in settings with limited visual information, where geometric relationships must be inferred from incomplete visual cues. Furthermore, a fine-grained analysis of MathVision reveals notable improvements in tasks that often benefit from auxiliary constructions, including Angle (+8.5), Length (+5.7), and Area (+4.6). These findings suggest that AuxCA can generalize learned construction strategies, such as drawing altitudes and extending line segments, to diverse geometric reasoning tasks, potentially improving reasoning robustness and accuracy beyond the training domain. We also evaluate AuxCA-8B on problems that do not require additional auxiliary constructions, as shown in Appendix~\ref{sec:no_aux}.

% \subsection{Generalization Analysis}

% To further evaluate the generalization ability of AuxCA, we conduct zero-shot evaluations on MathVista, MathVerse, and MathVision, as summarized in Table~\ref{tab:generalization_final_fix}. The results show that AuxCA consistently outperforms the base model across all three benchmarks. Notably, on the Text Dominant subset of MathVerse, AuxCA achieves a 7.2-point improvement, substantially exceeding the gain observed on the Text Lite subset. This suggests that the auxiliary-construction strategies learned during training may also facilitate reasoning in settings with limited visual information. Furthermore, a fine-grained analysis of MathVision reveals notable improvements in geometry-related categories that often benefit from auxiliary constructions, including Angle (+8.5), Length (+5.7), and Area (+4.6). These findings suggest that AuxCA learns transferable geometric reasoning strategies, such as drawing altitudes and extending line segments, that generalize beyond the specific construction patterns encountered during training. Such capabilities may contribute to more effective reasoning across diverse geometric problems. We additionally evaluate AuxCA-8B on problems that do not require auxiliary constructions, with results provided in Appendix~\ref{sec:no_aux}.

\subsection{Comparison with Generic Credit Assignment Methods}

\begin{wraptable}{r}{0.48\linewidth}
\vspace{-8pt}
\centering
\caption{\textbf{Comparison with Generic Credit Assignment Methods.}}
\label{table:generric_ca}

% \vspace{-1pt}
\small
\renewcommand{\arraystretch}{1.15}
\setlength{\tabcolsep}{5pt}

\begin{tabular}{lcc}
\toprule
\multirow{2}{*}{Method} & \multicolumn{2}{c}{Aux-PlaGeo} \\
\cmidrule(lr){2-3}
 & Complete & Weighted \\
\midrule
AuxCA              & \textbf{31.7} & \textbf{44.6} \\
VICRA   & 24.8 & 37.2 \\
HICRA   & 24.3 & 36.1 \\
TR-GRPO   & 18.9 & 30.2 \\
Tw-GRPO   & 17.6 & 28.7\\
\bottomrule
\end{tabular}
\vspace{-10pt}
\end{wraptable}

We conducted additional experiments comparing ACD with four generic credit assignment methods, as shown in Table~\ref{table:generric_ca}. We observe that VICRA, which amplifies the advantages of image-related tokens, and HICRA, which amplifies the advantages of planning-related reasoning tokens, both achieve slight improvements over SFT, but still fall far short of AuxCA. In contrast, TR-GRPO, which assigns larger weights to higher-probability tokens, and TW-GRPO, which places greater emphasis on high-entropy tokens, both show performance degradation relative to SFT. This indicates that our mechanism, which reallocates token-level advantages based on the difference in output token probabilities with and without auxiliary clues, is better suited to geometry problem solving than these more generic credit assignment methods.

\subsection{Case Study}

\begin{figure*}
  % \vspace{-5pt} % (可选) 微调垂直位置，有时需要向上移动一点
  \centering % 图片居中
  \includegraphics[width=1\textwidth]{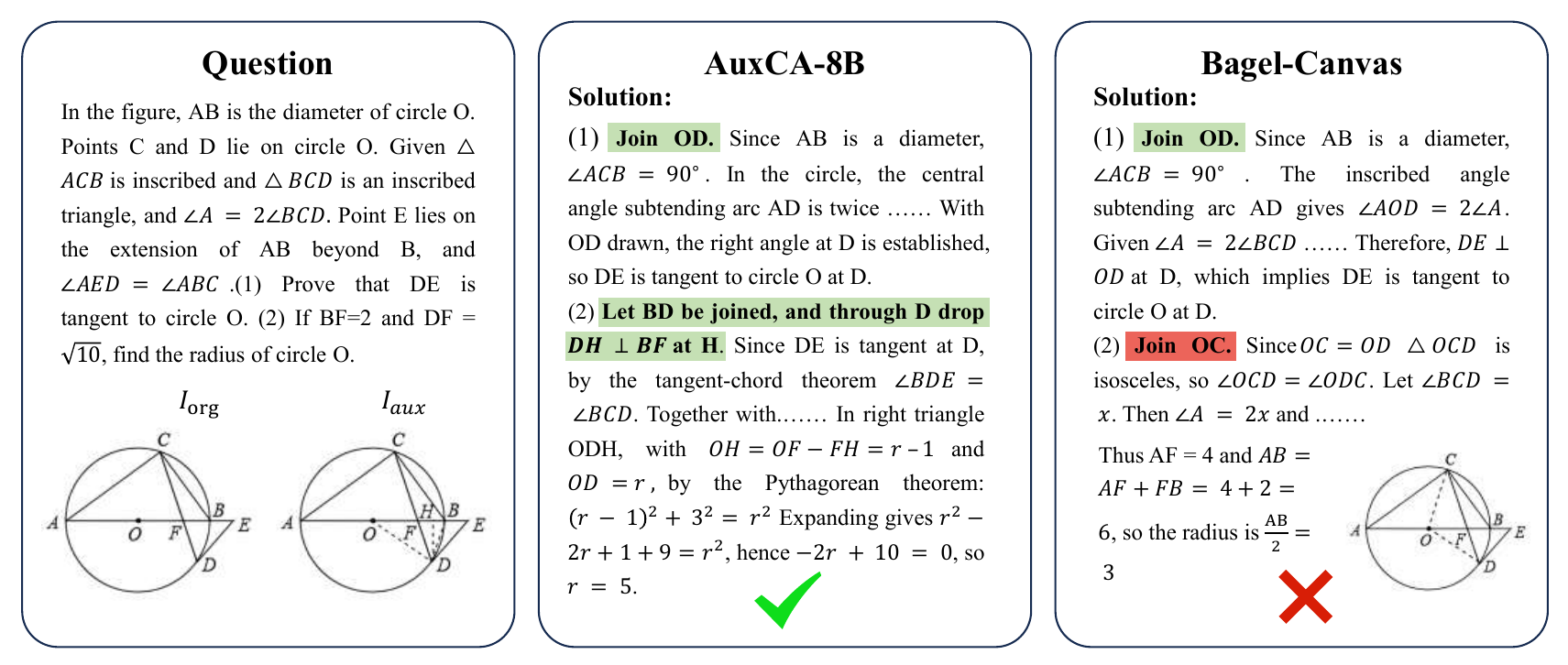}
\vspace{-10pt}
\caption{
\textbf{Performance comparison with Bagel-Canvas on a complex geometric circle problem.}}
\label{fig:case}
\vspace{-10pt}
\end{figure*}

To illustrate AuxCA's superiority in handling complex auxiliary constructions, Figure~\ref{fig:case} presents a representative circle geometry problem. In the foundational first part, both AuxCA and the explicit construction model, Bagel-Canvas, accurately identified the key clue (connecting OD) to complete the proof, confirming that both possess basic geometric perception. The performance gap widens in the challenging second part, which requires two non-intuitive auxiliary lines (connecting BD and dropping $DH \perp BF$) to form a right triangle. Despite generating no image, AuxCA precisely described this step in its reasoning text "Let BD be joined, and through D drop $DH \perp BF$" and correctly derived $r=5$. This example illustrates how AuxCA can propose useful auxiliary constructions through textual reasoning without generating additional diagrams. In contrast, Bagel-Canvas fell into a visual trap by explicitly drawing an ineffective line (OC). This misleading visual feedback caused "Error Cascading" and a calculation failure. Ultimately, this demonstrates that AuxCA avoids the pitfalls of explicit plotting, exhibiting more robust geometric structuring capabilities than explicit VCoT in complex, non-visual scenarios. In addition, we also analyze the failure cases of AuxCA-8B, with detailed results provided in the Appendix~\ref{sec:failure_analysis}.

% \section{Conclusion}

% To address the bottleneck of auxiliary construction in plane geometry reasoning, we present AuxCA, a novel reinforcement learning framework. Unlike traditional paradigms that rely on external plotting tools, AuxCA aims to internalize the ability to construct geometric clues within the model’s intrinsic reasoning capability. This is achieved through a synergistic mechanism that combines Auxiliary Clues Dependency (ACD) for fine-grained advantage weighting with a targeted auxiliary incentive reward. Extensive experiments demonstrate the superiority of our method, showing that implicit internalization offers a robust path toward solving complex geometric problems.

\section{Conclusion}

We present AuxCA, a reinforcement learning framework that helps models learn transferable auxiliary-construction strategies for geometric reasoning without explicit plotting at inference time. By combining Auxiliary Clues Dependency (ACD) for fine-grained advantage weighting with an auxiliary incentive reward, AuxCA improves problem-solving performance. Extensive experiments demonstrate its effectiveness and generalization beyond the plane geometry training domain.

\section*{Acknowledgments}
This work was supported by the "Pioneer" and "Leading Goose" R\&D Program of Zhejiang (Grant No. 2025C02022), the Ningbo "Yongjiang Talent Program" Youth Innovation Project (Grant No. 2024A-156-G), and the Young Scientists Fund of the National Natural Science Foundation of China (Grant No. 62507040). 

\bibliographystyle{plain} 
\bibliography{references}

%%%%%%%%%%%%%%%%%%%%%%%%%%%%%%%%%%%%%%%%%%%%%%%%%%%%%%%%%%%%
\clearpage
\appendix

\lstset{
  basicstyle=\small,   % 设置字体大小
  breaklines=true,              % 自动换行
  lineskip=1pt,                 % 行间距
    columns=fullflexible
}
\newtcblisting{mylisting}{
  listing only,
  colback=gray!5,               % 背景色
  colframe=black!50,            % 边框色
  arc=3mm,                      % 圆角半径
  boxrule=0.5pt,                % 边框粗细
  left=4pt, right=0pt, top=1pt, bottom=1pt, % 内边距
  listing options={
  % basicstyle=\normalsize
  basicstyle=\small, 
    % breaklines=true,            % 自动换行
    % lineskip=1pt,               % 行间距
    % columns=fullflexible        % 避免字间距过大
  }
}

\section{Prompts}
\label{prompt}
\subsection{Prompts used to reward model during training}
\label{sec:rewardprompt}
\begin{mylisting}
You are an expert mathematics teacher and a precise data evaluator.
Your task is to analyze a given math problem, compare a predicted solution against a ground truth answer, and determine if the prediction is correct.
Additionally, you need to assess whether the predicted solution relies on visual cues or supplementary elements present in the image but not explicitly specified in the problem text.

**INPUT FORMAT:**
You will be provided with a JSON string containing the following fields:
- `question_text`: The full text of the mathematical problem.
- `gt_answers`: A LIST of the final answers' strings (e.g., ["5 cm", "10 cm"]). Each element is the correct, final answer for a specific sub-question. This is the gold standard and no additional extraction is required.
- `prediction_solution`: The full solution text from the model, from which you must extract the final answer(s).

**TASK & OUTPUT REQUIREMENTS:**
Your output must be a single, valid JSON object.
The process involves three main steps: **Answer Parsing**, **Correctness Judgment**, and **Auxiliary Element Detection**.
---

### **Step 1: Answer Parsing and Extraction**
Your task is to create a list of predicted answers (named `pred_answers`) based on the given `prediction_solution`. This list must match the provided list of correct answers (`gt_answers`) exactly.
**Parsing and Extraction Rules:**
- Your primary task is to **extract** the final answer(s) from the `prediction_solution` text to create the `pred_answers` list.
- **IMPORTANT:** The answers to different sub-questions may appear in different places within the `prediction_solution`, not necessarily grouped together at the end. You must treat this as a **matching task**.
- For each part of the `gt_answers` list, you must scan the **entire** `prediction_solution` to find the corresponding predicted answer. Look for explicit labels (e.g., "(1)", "Part A"), final conclusions, boxed answers (e.g., \\boxed{{...}}), or statements that directly answer a part of the original question.
- The final `pred_answers` list **must have the exact same length as the `gt_answers` list**.
- For each sub-question, if you cannot find a corresponding answer in the `prediction_solution`, you **must** use `null` as a placeholder in that position.
**This rule is critical and applies in all cases where an answer is missing, including when the `prediction_solution` appears incomplete or is truncated before all sub-questions are addressed.**
**CRITICAL RULE:** The final `gt_answers` and `pred_answers` lists **must** be of equal length. The number of parts in `gt_answers` dictates the required length.

### **Step 2: Correctness Judgment**
Your second task is to compare the `pred_answers` list against the `gt_answers` list, element by element.
**Judgment Rules:**
1. **Numerical Equivalence:** Treat numbers as correct if they are mathematically equivalent (e.g., `5`, `5.0`, `10/2`). Allow for minor floating-point rounding differences.
2. **Textual Equivalence:** For text answers, judge based on semantic meaning, not exact matching. Ignore case, whitespace, and phrasing differences (e.g., "CB is parallel to PD" is equivalent to "Line CB || Line PD").
3. **Generate Correctness List:** Create a boolean list named `correctness`. The i-th element is `true` if the i-th predicted answer is correct, `false` otherwise. This list **must** have the same length as the answer lists.
\end{mylisting}

\begin{mylisting}
### **Step 3: Detection of Pre-drawn Auxiliary Elements**
The input image for this problem contains pre-drawn auxiliary elements (i.e., additional visual elements not explicitly mentioned in the problem description, such as points, lines, circles, etc.), intended to assist in solving the problem.
Analyze the reasoning process of `prediction_solution` to determine whether it utilizes any pre-drawn auxiliary elements visible in the image.
Since the image cannot be viewed, **infer from the text**:
- Does the solution reference additional visual elements not explicitly mentioned in the problem description, such as points, lines, circles, etc.?
- Does the solution rely on specific geometric relationships (e.g., similar triangles or the Pythagorean theorem) that require constructions implied by the image but not explicitly stated in `question_text`?
- **Evaluation Criteria:** Set `using_auxiliary` to `true` if the solution explicitly references these auxiliary elements or their derived geometric properties not mentioned in the problem text. Otherwise, set it to `false`.
---

**Final JSON Output Structure:**
Your entire response must be a single, valid JSON object matching the schema below. Do not include any text outside of this JSON object.
```json
{{
  "using_auxiliary": true/false,
  "correctness": [
    true/false,
    ...
  ]
}}
```
**INPUT DATA:**
f{input_data}
Process this mathematical problem according to all requirements above. Ensure the output is valid JSON.
\end{mylisting}

\subsection{Prompts used to data selection}
\label{sec:dataselect}
\label{sec:appendix_data_construction}

\begin{mylisting}
I will present you with two geometric images. The first is the [Original Problem Image], and the second is the [Auxiliary Image].
Please observe these two images carefully.
Task: Determine whether the [Auxiliary Image] is derived merely by adding auxiliary lines (such as dashed lines, red lines, extension lines, etc.) or other letter/character markings to the [Original Problem Image], while the shape and position of the main geometric subject remain unchanged.
If yes, please answer only: YES
If the [Auxiliary Image] is a completely different image, or differs significantly from the [Original Problem Image], please answer: NO
After your reasoning, please start a new line and **output only the conclusion string ("<|begin_of_box|>YES<|end_of_box|>" or <|begin_of_box|>NO<|end_of_box|>")**. That is:
If it is an auxiliary line relationship, please output: <|begin_of_box|>YES<|end_of_box|>
If not, please output: <|begin_of_box|>NO<|end_of_box|>"
\end{mylisting}

\section{Automated Filtering Pipeline}
\label{sec:data_pipeline}

To construct a dedicated dataset for auxiliary-dependent plane geometry, we performed a rigorous filtering process on the Plane Geometry subset of \textit{MathCanvas-Instruct}. Our objective was to isolate problems where the solution strictly necessitates the addition of auxiliary geometric structures (e.g., auxiliary lines, extended segments) rather than generating entirely new figures or solving via pure calculation.

Given the scale of the dataset, manual verification was infeasible. Therefore, we developed an automated pipeline leveraging the advanced visual reasoning capabilities of \textbf{GLM-4.5V}. For each data sample, we fed the model a pair of images:
\begin{itemize}
    \item \textbf{Original Image ($I_{\text{org}}$):} The initial geometric figure provided in the problem statement.
    \item \textbf{Auxiliary Image ($I_{\text{aux}}$):} The figure parsed from the solution steps, containing potential auxiliary constructions.
\end{itemize}

The VLM was tasked with verifying whether $I_{\text{aux}}$ is visually derived from $I_{\text{org}}$ solely by adding auxiliary markers without altering the position or shape of the main geometric subjects. 

We employed a strict prompt showing in Appendix~\ref{sec:appendix_data_construction} template to ensure high precision in data selection. Only samples where the model returned the token \texttt{YES} were included in our final \textbf{Aux-PlaGeo} training set.

\section{Proof of Total Advantage Conservation}
\label{sec:proof_conservation}

In this section, we provide the detailed mathematical derivation for the \textit{Total Advantage Conservation} property presented in Eq.~(6) of the main paper. This property ensures that our reweighting mechanism redistributes the advantage within a sequence based on auxiliary dependency without altering the cumulative advantage of the entire sequence, thereby maintaining training stability.

\subsection{Definitions and Preliminaries}

Recall the definition of the token-level weighting function $\psi(d_{i,t})$ from Eq.~(5):
\begin{equation}
    \psi(d_{i,t}) = 1 + \left( d_{i,t} - \frac{1}{|y_i|} \sum_{k=1}^{|y_i|} d_{i,k} \right),
    \label{eq:appendix_psi_def}
\end{equation}
where $y_i$ represents the sampled sequence, $|y_i|$ denotes its length, and $d_{i,t}$ is the auxiliary dependency for the $t$-th token.

In the GRPO framework, the advantage $A_{i,t}$ is typically derived from the sequence-level reward. Therefore, for a given sequence $y_i$, the advantage value is constant across all tokens:
\begin{equation}
    A_{i,t} = A_i, \quad \forall t \in \{1, \dots, |y_i|\}.
\end{equation}

\subsection{Derivation}

We aim to prove that the total advantage remains unchanged after reweighting:
\begin{equation}
    \sum_{t=1}^{|y_i|} A_{i,t} = \sum_{t=1}^{|y_i|} \psi(d_{i,t}) \cdot A_{i,t}.
\end{equation}

\textit{Proof.} Starting from the right-hand side (RHS) of the equation:

\begin{align*}
    \text{RHS} &= \sum_{t=1}^{|y_i|} \psi(d_{i,t}) \cdot A_{i,t} \\
    &= \sum_{t=1}^{|y_i|} \left[ 1 + \left( d_{i,t} - \frac{1}{|y_i|} \sum_{k=1}^{|y_i|} d_{i,k} \right) \right] \cdot A_i \quad \text{(Substitute } \psi \text{ and } A_{i,t}=A_i \text{)} \\
    &= A_i \cdot \sum_{t=1}^{|y_i|} \left( 1 + d_{i,t} - \bar{d}_i \right) \quad \text{(Let } \bar{d}_i = \frac{1}{|y_i|} \sum_{k=1}^{|y_i|} d_{i,k} \text{)} \\
    &= A_i \cdot \left( \sum_{t=1}^{|y_i|} 1 + \sum_{t=1}^{|y_i|} d_{i,t} - \sum_{t=1}^{|y_i|} \bar{d}_i \right) \quad \text{(Linearity of summation)} \\
    &= A_i \cdot \left( |y_i| + |y_i| \cdot \bar{d}_i - |y_i| \cdot \bar{d}_i \right) \quad \text{(Note that } \sum d_{i,t} = |y_i|\bar{d}_i \text{)} \\
    &= A_i \cdot |y_i| \\
    &= \sum_{t=1}^{|y_i|} A_i \\
    &= \sum_{t=1}^{|y_i|} A_{i,t} = \text{LHS}.
\end{align*}

\hfill $\square$

\subsection{Remark}
The derivation above confirms that the term $\left( d_{i,t} - \bar{d}_i \right)$ acts as a zero-sum adjustment term. This guarantees that while tokens with higher auxiliary dependency ($d_{i,t} > \bar{d}_i$) receive amplified advantages, and those with lower dependency receive attenuated advantages, the total policy gradient magnitude for the sequence remains consistent with the original GRPO baseline.

\section{Ablation Study on the Mapping Function $\psi$}
\label{sec:appendix_psi_ablation}

To determine the optimal strategy for converting the auxiliary dependency into advantage weights, we investigated two distinct mapping functions during the exploration phase on the \textbf{Qwen2.5-VL-3B-Instruct} model. Let $d_{i,t}$ denote the token-level auxiliary dependency and $\bar{d}_i$ be the sequence mean.

\subsection{Candidate Functions}

We compared a linear centering approach against a non-linear exponential approach:

\begin{itemize}
    \item \textbf{Linear Mapping ($\psi_1$):} This function applies a centered linear adjustment:
    \begin{equation}
        \psi_1(d_{i,t}) = 1 + \left( d_{i,t} - \frac{1}{|y_i|} \sum_{k=1}^{|y_i|} d_{i,k} \right).
    \end{equation}
    
    \item \textbf{Exponential Mapping ($\psi_2$):} This function applies a Softmax-like non-linear scaling:
    \begin{equation}
        \psi_2(d_{i,t}) = \exp \left( d_{i,t} - \frac{1}{|y_i|} \sum_{k=1}^{|y_i|} d_{i,k} \right).
    \end{equation}
\end{itemize}

\subsection{Results and Analysis}

The comparative results are presented in Table~\ref{tab:psi_ablation}.

\begin{table}[h]
    \centering
    \caption{\textbf{Comparison of different mapping functions on Qwen2.5-VL-3B-Instruct.} The Linear Mapping ($\psi_1$) outperforms the Exponential Mapping ($\psi_2$).}
    \label{tab:psi_ablation}
    \setlength{\tabcolsep}{20pt}
    \renewcommand{\arraystretch}{1.3}
    \begin{tabular}{lcc}
        \toprule
        \textbf{Function} & \textbf{Formulation} & \textbf{Complete Score} \\
        \midrule
        $\psi_1$ (Linear) & $1 + (d_{i,t} - \bar{d}_i)$ & \textbf{7.15\%} \\
        $\psi_2$ (Exponential) & $\exp(d_{i,t} - \bar{d}_i)$ & 6.30\% \\
        \bottomrule
    \end{tabular}
\end{table}

\textbf{Analysis.} The experimental evidence favors the Linear Mapping ($\psi_1$), which yields a performance gain of \textbf{+0.85\%} compared to $\psi_2$. 

From a theoretical perspective, this result aligns with the \textit{Total Advantage Conservation} property (proved in Sec.~\ref{sec:proof_conservation}). The linear function $\psi_1$ ensures that the sum of advantages remains invariant after re-weighting ($\sum \psi_1 \cdot A = \sum A$), acting purely as an internal redistribution mechanism. Conversely, the exponential function $\psi_2$ distorts the total advantage magnitude of the sequence. This distortion likely introduces variance and instability into the gradient estimation, leading to suboptimal convergence. 

Based on these findings, we selected $\psi_1$ as the standard mapping function for training the larger \textbf{Qwen3-VL-Thinking} model.

\section{Selection of Auxiliary Incentive Reward $r_{\text{aux}}$}
\label{sec:appendix_hyperparam_3}

To determine the optimal magnitude for the auxiliary incentive reward $r_{\text{aux}}$, we conducted a pilot study using the \textbf{Qwen2.5-VL-3B-Instruct} model. The coefficient $r_{\text{aux}}$ plays a pivotal role in balancing the model's focus between primary reasoning and auxiliary structure utilization. We evaluated the model's Complete Score across a range of values from $0$ to $0.5$ on the overall MathCanvas-Bench. The results are summarized in Table~\ref{tab:raux_ablation}.

\begin{table}[h]
    \centering
    \caption{\textbf{Sensitivity analysis of $r_{\text{aux}}$ on Qwen2.5-VL-3B-Instruct.} We swept the auxiliary reward weight from 0 to 0.5. The best performance is highlighted in \textbf{bold}.}
    \label{tab:raux_ablation}
    \setlength{\tabcolsep}{12pt} % 增加列间距，使表格更美观
    \renewcommand{\arraystretch}{1.2} % 增加行高
    \begin{tabular}{lcccccc}
        \toprule
        \textbf{$r_{\text{aux}}$ Value} & 0 & 0.1 & \textbf{0.2} & 0.3 & 0.4 & 0.5 \\
        \midrule
        \textbf{Complete Score} & 7.15\% & 7.36\% & \textbf{7.50\%} & 7.15\% & 7.28\% & 6.93\% \\
        \bottomrule
    \end{tabular}
\end{table}

\textbf{Analysis.} As illustrated in Table~\ref{tab:raux_ablation}, the performance exhibits an inverted U-shaped trend. Initially, increasing $r_{\text{aux}}$ yields performance gains, peaking at \textbf{7.50\%} when $r_{\text{aux}}=0.2$. This indicates that a moderate auxiliary signal effectively encourages the model to leverage geometric clues. However, performance degrades as $r_{\text{aux}}$ increases further ($>0.2$), likely because an excessive weight on the auxiliary objective interferes with the optimization of the primary reasoning reward ($r_{\text{acc}}$).

\textbf{Decision.} Based on this pilot exploration, we selected $r_{\text{aux}}=0.2$ as the optimal hyperparameter. We applied this setting to the training of the main \textbf{Qwen-3-VL-Thinking} model to support the learning of transferable auxiliary-construction strategies.

\section{Implementation Details}
\label{sec:appendix_implementation}

\subsection{details on the LLM judge}

We employ DeepSeek-V3.2 as our LLM judge (prompt in Appendix A.1). At the early stage of our study, to verify its reliability, we compared it with human evaluation and other judge models on 100 randomly sampled instances from Aux-PlaGeo, repeating five times, with results provided in Table~\ref{tab:different_judges}. All models achieved over 98\% consistency with human evaluation owing to reference answers. Considering efficiency and cost, we selected DeepSeek-V3.2 as the judge model.

\begin{table}[t]
    \centering
    \caption{Comparison of Different Judges}
    \label{tab:different_judges}
    \small
    \renewcommand{\arraystretch}{1.2}
    \begin{tabular}{lcccccc}
        \toprule
        \textbf{Judge} & \textbf{1} & \textbf{2} & \textbf{3} & \textbf{4} & \textbf{5} & \textbf{Avg} \\
        \midrule
        DeepSeek-V3.2 & 99 & 98 & 99 & 100 & 98 & 98.8\% \\
        GLM-4.6      & 98 & 99 & 98 & 99  & 97 & 98.2\% \\
        GPT-5        & 99 & 99 & 100 & 98 & 99 & \textbf{99.0\%} \\
        \bottomrule
    \end{tabular}
\end{table}

\subsection{Impact of training sample size}

We experimented with various RL dataset sizes after one training epoch. As shown in Table~\ref{tab:training_sample_size}, performance rises steadily with larger datasets but shows diminishing marginal gains: expanding from 14k to 15k only improves complete accuracy by 0.5\%. The 15k setting achieves the best overall performance at acceptable cost, so we use it as our main configuration.

\begin{table}
    \centering
    \caption{Impact of training sample size}
    \label{tab:training_sample_size}
    \small
    \renewcommand{\arraystretch}{1.2}
    \begin{tabular}{lcc}
        \toprule
        \textbf{Size} & \textbf{Complete} & \textbf{Weighted} \\
        \midrule
        5k  & 24.1 & 34.7 \\
        10k & 28.4 & 40.3 \\
        12k & 29.9 & 42.4 \\
        14k & 31.2 & 43.9 \\
        15k & \textbf{31.7} & \textbf{44.6} \\
        \bottomrule
    \end{tabular}
\end{table}

\subsection{Hyperparameter Settings}

We provide the detailed hyperparameters used during the training phase in Table~\ref{tab:hyperparameters}. The experiments were conducted using a computational node equipped with 8 NVIDIA A100 GPUs. 

We utilized the Hugging Face \texttt{trl} library to implement the training pipeline. For reproducibility, please note that \textbf{any hyperparameters not explicitly listed in Table~\ref{tab:hyperparameters} follow the default settings of the \texttt{trl} library}.

\begin{table}[h]
    % ---> 核心修改：强制顶端对齐 <---
    \makeatletter
    \setlength{\@fptop}{0pt}
    \makeatother
    % ---------------------------
    
    \centering
    \caption{\textbf{Hyperparameters used for training AuxCA.} The model is trained using full fine-tuning (without PEFT) on 8 $\times$ A100 GPUs.}
    \label{tab:hyperparameters}
    \setlength{\tabcolsep}{18pt} 
    \renewcommand{\arraystretch}{1.2} 
    \begin{tabular}{lc}
        \toprule
        \textbf{Parameter} & \textbf{Value} \\
        \midrule
        Learning Rate (LR) & 2e-6 \\
        LR Scheduler & Cosine Decay \\
        Warmup Ratio & 0.1 \\
        Use PEFT & False \\
        Rollout & 8 \\
        Per-Device Train Batch Size & 2 \\
        Max Completion Length & 4096 \\
        Max Prompt Length & 1024 \\
        Gradient Accumulation Steps & 4 \\
        Num Train Epochs & 1 \\
        Computing Infrastructure & 8 $\times$ A100 \\
        \bottomrule
    \end{tabular}
\end{table}

\section{Performance on problems that do not require auxiliary construction}
\label{sec:no_aux}

We conducted an additional experiment using the manually extracted subset of geometry problems in MathVista testmini that require no auxiliary construction, and compared AuxCA with the base model. As shown in Table~\ref{tab:mathvista_no_aux}, AuxCA outperforms the base model by only 0.9\% on the w/o aux subset, verifying that AuxCA training preserves the model’s original reasoning ability while bringing a minor gain. In contrast, it yields a 2.9\% improvement on the overall set, demonstrating significant performance boosts on problems requiring auxiliary construction as well as strong zero-shot reasoning ability on other benchmarks.

\begin{table}[t]
    \centering
    \caption{Performance on the MathVista}
    \label{tab:mathvista_no_aux}
    \small
    \renewcommand{\arraystretch}{1.2}
    \begin{tabular}{lcc}
        \toprule
        \textbf{Model} & \textbf{GPS} & \textbf{GPS w/o aux} \\
        \midrule
        AuxCA & \textbf{71.6} & \textbf{73.7} \\
        base model & 68.7 & 72.8 \\
        \bottomrule
    \end{tabular}
\end{table}

\section{Analysis on failure cases}
\label{sec:failure_analysis}

\subsection{Failure Analysis}
We have added a failure case analysis. Using five error categories defined in Table~\ref{tab:failure_analysis}, we employed DeepSeek\mbox{-}V3.2 to classify incorrect model outputs based on the problem and ground truth. As shown in Table~\ref{tab:failure_analysis}, 70.6\% of errors stem from "correct clue but wrong reasoning", meaning AuxCA helps the model construct auxiliary elements effectively, yet its reasoning with such clues is still insufficient. We will include a detailed failure case study in the paper and investigate how to strengthen this reasoning ability in future work.

\begin{table}
    \centering
    \caption{Failure Analysis}
    \label{tab:failure_analysis}
    \small
    \renewcommand{\arraystretch}{1.2}
    \begin{tabular}{lc}
        \toprule
        \textbf{Error Type} & \textbf{\%} \\
        \midrule
        Correct clue but wrong reasoning & \textbf{70.6} \\
        Construct wrong auxiliary clue & 12.5 \\
        Incomplete solution & 8.3 \\
        Missing auxiliary construction & 3.3 \\
        Other & 5.3 \\
        \bottomrule
    \end{tabular}
\end{table}

\section{Inference Cost Comparison with Explicit Plotting}
\label{sec:inference_cost}

\begin{table}[t]
\centering
\caption{Inference cost on 635 \texttt{Aux-PlaGeo} problems under the same hardware and inference settings. Token counts include text output only; latency covers the complete inference process.}
\label{tab:inference_cost}
\small
\begin{tabular}{lcc}
\toprule
Model & Avg. text tokens & Avg. latency (s) \\
\midrule
AuxCA-8B & 1,723 & \textbf{17.38} \\
Bagel-Canvas & 683 & 50.24 \\
\bottomrule
\end{tabular}
\end{table}

We compare AuxCA-8B and Bagel-Canvas on all 635 \texttt{Aux-PlaGeo} problems under the same hardware and inference settings. As shown in Table~\ref{tab:inference_cost}, AuxCA generates 1,723 text tokens per problem on average, compared with 683 for Bagel-Canvas. Nevertheless, its average end-to-end latency is 17.38 seconds, compared with 50.24 seconds for Bagel-Canvas, representing a 65.4\% reduction. Text token counts do not capture the work required to generate images, whereas end-to-end latency includes the complete inference process. Together with the accuracy gains reported in Table~\ref{table:main}, these results show that AuxCA achieves higher accuracy and lower latency on this subset despite producing longer textual responses.

\section{Analysis of ACD Scores}
\label{sec:acd_scores}

\begin{wrapfigure}{r}{0.60\linewidth}
    \vspace{-32pt}
    \centering
    \includegraphics[width=\linewidth]{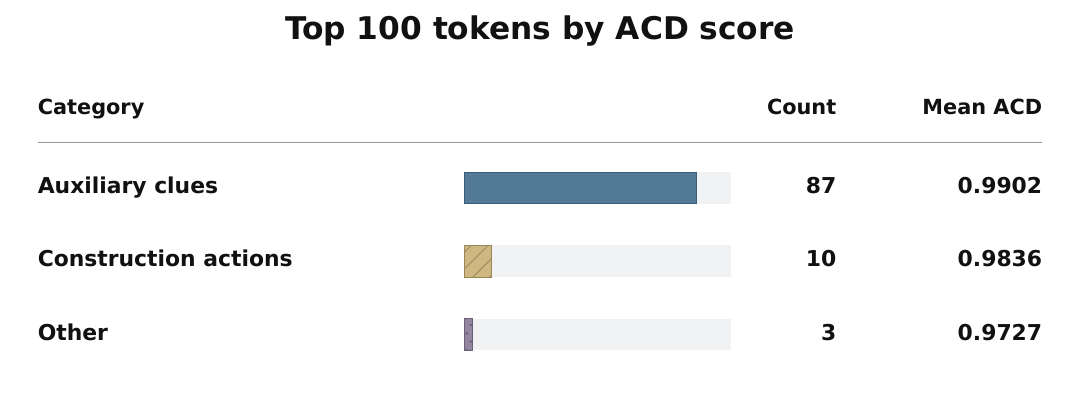}
    \setlength{\abovecaptionskip}{2pt}
    \vspace{-10pt}
    \caption{\textbf{Top 100 tokens by ACD.}}
    \label{fig:acd_top_tokens}
    \vspace{-16pt}
\end{wrapfigure}

To examine whether ACD identifies auxiliary-construction content, we randomly sample 200 \texttt{Aux-PlaGeo} test problems and compute ACD for every generated token.

As shown in Figure~\ref{fig:acd_top_tokens}, 87 of the 100 highest-scoring tokens are manually identified as auxiliary-clue tokens, such as ``D'' and ``DE'', with a mean ACD of 0.9902. Another 10 are construction-action tokens, including ``connect'', ``extend'', and ``join'', with a mean ACD of 0.9836; only 3 belong to other categories. The prominence of both geometric elements and construction actions suggests that ACD captures tokens relevant to auxiliary construction. We further examine low-scoring tokens and the overall score distribution in Appendix~\ref{sec:acd_distribution}. These analyses reveal a clear correspondence between ACD scores and auxiliary-construction content identified by human annotators, demonstrating the effectiveness of ACD.

\section{Further Analysis of ACD Scores}
\label{sec:acd_distribution}

We further analyze the ACD scores of tokens generated on the 200 sampled \texttt{Aux-PlaGeo} problems. The 100 lowest-scoring tokens have a mean ACD of $-0.4631$ and are mainly function words, such as ``the'', ``a'', and ``and''. Unlike the auxiliary-clue and construction-action tokens concentrated at the high end of the ranking, these tokens generally do not express geometric construction steps.

Table~\ref{tab:acd_distribution} reports the distribution of scores over all generated tokens. Positive scores account for 34.2\% of tokens, negative scores for 20.7\%, and recorded zero scores for 45.1\%. The zero-score group consists mainly of punctuation, numbers, and mathematical operators, for which no probability shift between the auxiliary-enhanced and original images was recorded. Under the centered weighting function $\psi(d_{i,t})=1+(d_{i,t}-\bar d_i)$, tokens with higher ACD scores receive larger relative weights within a response; a zero ACD score does not itself imply zero training weight. Together with the manually annotated high-scoring tokens in Figure~\ref{fig:acd_top_tokens}, this distribution supports the intended role of ACD in distinguishing auxiliary-construction content from more generic tokens.

\begin{table}[t]
    \centering
    \caption{\textbf{Distribution of token-level ACD scores.}
    Percentages are computed over all generated tokens from the 200 sampled \texttt{Aux-PlaGeo} problems.}
    \label{tab:acd_distribution}
    \small
    \begin{tabular}{lc}
        \toprule
        ACD score & Percentage \\
        \midrule
        $d_t > 0$ & 34.2\% \\
        $d_t = 0$ & 45.1\% \\
        $d_t < 0$ & 20.7\% \\
        \bottomrule
    \end{tabular}
\end{table}

\section{Limitations.}
Despite the effectiveness of AuxCA in encouraging auxiliary clue construction, our failure analysis indicates that this ability does not always translate into correct reasoning. As shown in Table~\ref{tab:failure_analysis}, 70.6\% of the errors fall into the ``correct clue but wrong reasoning'' category, suggesting that the model can often construct useful auxiliary elements but still struggles to reliably exploit them in multi-step geometric reasoning. This highlights reasoning over constructed clues as a key bottleneck of the current framework. In addition, our error categorization is based on automatic classification by DeepSeek-V3.2, which may introduce annotation noise for ambiguous or partially correct solutions. Future work will investigate stronger reasoning mechanisms and more rigorous verification strategies to better leverage correctly constructed auxiliary clues.

\section{Social Impact.}
The proposed method may have positive social impacts by improving AI-assisted education and visual reasoning. In particular, AuxCA can help models better identify and utilize auxiliary geometric structures, which may support intelligent tutoring systems, provide more interpretable problem-solving feedback, and help students understand complex geometric reasoning processes.

At the same time, the method also poses potential risks. More capable reasoning models may be misused for academic dishonesty if students rely on them to directly obtain solutions. Moreover, models may still generate plausible but incorrect reasoning, which could mislead users if not properly verified. Therefore, AuxCA-based systems should be deployed as assistive tools with appropriate safeguards, transparent reasoning outputs, and human oversight, especially in educational or high-stakes scenarios.

%%%%%%%%%%%%%%%%%%%%%%%%%%%%%%%%%%%%%%%%%%%%%%%%%%%%%%%%%%%%

\newpage
\section*{NeurIPS Paper Checklist}

%%% BEGIN INSTRUCTIONS %%%
The checklist is designed to encourage best practices for responsible machine learning research, addressing issues of reproducibility, transparency, research ethics, and societal impact. Do not remove the checklist: {\bf The papers not including the checklist will be desk rejected.} The checklist should follow the references and follow the (optional) supplemental material.  The checklist does NOT count towards the page
limit. 

Please read the checklist guidelines carefully for information on how to answer these questions. For each question in the checklist:
\begin{itemize}
    \item You should answer \answerYes{}, \answerNo{}, or \answerNA{}.
    \item \answerNA{} means either that the question is Not Applicable for that particular paper or the relevant information is Not Available.
    \item Please provide a short (1--2 sentence) justification right after your answer (even for \answerNA). 
   % \item {\bf The papers not including the checklist will be desk rejected.}
\end{itemize}

{\bf The checklist answers are an integral part of your paper submission.} They are visible to the reviewers, area chairs, senior area chairs, and ethics reviewers. You will also be asked to include it (after eventual revisions) with the final version of your paper, and its final version will be published with the paper.

The reviewers of your paper will be asked to use the checklist as one of the factors in their evaluation. While \answerYes{} is generally preferable to \answerNo{}, it is perfectly acceptable to answer \answerNo{} provided a proper justification is given (e.g., error bars are not reported because it would be too computationally expensive'' or ``we were unable to find the license for the dataset we used''). In general, answering \answerNo{} or \answerNA{} is not grounds for rejection. While the questions are phrased in a binary way, we acknowledge that the true answer is often more nuanced, so please just use your best judgment and write a justification to elaborate. All supporting evidence can appear either in the main paper or the supplemental material, provided in appendix. If you answer \answerYes{} to a question, in the justification please point to the section(s) where related material for the question can be found.

IMPORTANT, please:
\begin{itemize}
    \item {\bf Delete this instruction block, but keep the section heading ``NeurIPS Paper Checklist"},
    \item  {\bf Keep the checklist subsection headings, questions/answers and guidelines below.}
    \item {\bf Do not modify the questions and only use the provided macros for your answers}.
\end{itemize}

%%% END INSTRUCTIONS %%%

\begin{enumerate}

\item {\bf Claims}
    \item[] Question: Do the main claims made in the abstract and introduction accurately reflect the paper's contributions and scope?
    \item[] Answer: \answerYes{} % Replace by \answerYes{}, \answerNo{}, or \answerNA{}.
    \item[] Justification: Main claims are clearly stated in Abstract and Introduction, consistent with theoretical/experimental results and scope in the paper.
    \begin{itemize}
        \item The answer \answerNA{} means that the abstract and introduction do not include the claims made in the paper.
        \item The abstract and/or introduction should clearly state the claims made, including the contributions made in the paper and important assumptions and limitations. A \answerNo{} or \answerNA{} answer to this question will not be perceived well by the reviewers. 
        \item The claims made should match theoretical and experimental results, and reflect how much the results can be expected to generalize to other settings. 
        \item It is fine to include aspirational goals as motivation as long as it is clear that these goals are not attained by the paper. 
    \end{itemize}

\item {\bf Limitations}
    \item[] Question: Does the paper discuss the limitations of the work performed by the authors?
    \item[] Answer: \answerYes{} % Replace by \answerYes{}, \answerNo{}, or \answerNA{}.
    \item[] Justification:  See Appendix H.

\item {\bf Theory assumptions and proofs}
    \item[] Question: For each theoretical result, does the paper provide the full set of assumptions and a complete (and correct) proof?
    \item[] Answer: \answerYes{} % Replace by \answerYes{}, \answerNo{}, or \answerNA{}.
    \item[] Justification: All theorems include clear assumptions; complete proofs are provided in the main text and Appendix.

    \item {\bf Experimental result reproducibility}
    \item[] Question: Does the paper fully disclose all the information needed to reproduce the main experimental results of the paper to the extent that it affects the main claims and/or conclusions of the paper (regardless of whether the code and data are provided or not)?
    \item[] Answer: \answerYes{} % Replace by \answerYes{}, \answerNo{}, or \answerNA{}.
    \item[] Justification: Full reproducibility details are disclosed in main text and Appendix, enabling replication of core experimental results.

\item {\bf Open access to data and code}
    \item[] Question: Does the paper provide open access to the data and code, with sufficient instructions to faithfully reproduce the main experimental results, as described in supplemental material?
    \item[] Answer: \answerYes{} % Replace by \answerYes{}, \answerNo{}, or \answerNA{}.
    \item[] Justification: Code and data are publicly accessible with detailed reproduction instructions in the main text and Appendix.

\item {\bf Experimental setting/details}
    \item[] Question: Does the paper specify all the training and test details (e.g., data splits, hyperparameters, how they were chosen, type of optimizer) necessary to understand the results?
    \item[] Answer: \answerYes{} % Replace by \answerYes{}, \answerNo{}, or \answerNA{}.
    \item[] Justification: Complete training/test details are provided in Section 4 and Appendix.

\item {\bf Experiment statistical significance}
    \item[] Question: Does the paper report error bars suitably and correctly defined or other appropriate information about the statistical significance of the experiments?
    \item[] Answer: \answerNo{} % Replace by \answerYes{}, \answerNo{}, or \answerNA{}.
    \item[] Justification: The reported evaluation results in this paper is averaged through multiple runs of experiments.

\item {\bf Experiments compute resources}
    \item[] Question: For each experiment, does the paper provide sufficient information on the computer resources (type of compute workers, memory, time of execution) needed to reproduce the experiments?
    \item[] Answer: \answerYes{} % Replace by \answerYes{}, \answerNo{}, or \answerNA{}.
    \item[] Justification: Compute resources are detailed for all experiments in Appendix F.
    
\item {\bf Code of ethics}
    \item[] Question: Does the research conducted in the paper conform, in every respect, with the NeurIPS Code of Ethics \url{https://neurips.cc/public/EthicsGuidelines}?
    \item[] Answer: \answerYes{} % Replace by \answerYes{}, \answerNo{}, or \answerNA{}.
    \item[] Justification: The entire research follows the NeurIPS Code of Ethics, with anonymity and ethical norms fully respected.

\item {\bf Broader impacts}
    \item[] Question: Does the paper discuss both potential positive societal impacts and negative societal impacts of the work performed?
    \item[] Answer: \answerYes{} % Replace by \answerYes{}, \answerNo{}, or \answerNA{}.
    \item[] Justification: Positive and negative societal impacts are analyzed in the Appendix H, with potential risks noted.
    
\item {\bf Safeguards}
    \item[] Question: Does the paper describe safeguards that have been put in place for responsible release of data or models that have a high risk for misuse (e.g., pre-trained language models, image generators, or scraped datasets)?
    \item[] Answer: \answerNA{} % Replace by \answerYes{}, \answerNo{}, or \answerNA{}.
    \item[] Justification: This paper does not involve the release of models or datasets with high misuse risk.

\item {\bf Licenses for existing assets}
    \item[] Question: Are the creators or original owners of assets (e.g., code, data, models), used in the paper, properly credited and are the license and terms of use explicitly mentioned and properly respected?
    \item[] Answer: \answerYes{} % Replace by \answerYes{}, \answerNo{}, or \answerNA{}.
    \item[] Justification: All data used in this study are obtained from publicly available datasets.  We have ensured that all licenses have been properly respected.

\item {\bf New assets}
    \item[] Question: Are new assets introduced in the paper well documented and is the documentation provided alongside the assets?
    \item[] Answer: \answerNA{} % Replace by \answerYes{}, \answerNo{}, or \answerNA{}.
    \item[] Justification: The paper does not release new assets.

\item {\bf Crowdsourcing and research with human subjects}
    \item[] Question: For crowdsourcing experiments and research with human subjects, does the paper include the full text of instructions given to participants and screenshots, if applicable, as well as details about compensation (if any)? 
    \item[] Answer: \answerNA{} % Replace by \answerYes{}, \answerNo{}, or \answerNA{}.
    \item[] Justification: This work does not involve crowdsourcing or research with human subjects.

\item {\bf Institutional review board (IRB) approvals or equivalent for research with human subjects}
    \item[] Question: Does the paper describe potential risks incurred by study participants, whether such risks were disclosed to the subjects, and whether Institutional Review Board (IRB) approvals (or an equivalent approval/review based on the requirements of your country or institution) were obtained?
    \item[] Answer: \answerNA{} % Replace by \answerYes{}, \answerNo{}, or \answerNA{}.
    \item[] Justification:  This work does not involve crowdsourcing or research with human subjects.

\item {\bf Declaration of LLM usage}
    \item[] Question: Does the paper describe the usage of LLMs if it is an important, original, or non-standard component of the core methods in this research? Note that if the LLM is used only for writing, editing, or formatting purposes and does \emph{not} impact the core methodology, scientific rigor, or originality of the research, declaration is not required.
    %this research? 
    \item[] Answer: \answerYes{} % Replace by \answerYes{}, \answerNo{}, or \answerNA{}.
    \item[] Justification: This paper uses LLM as a reward model, with its evaluation and consistency with humans reported in the Appendix F.

\end{enumerate}

\end{document}